# Commonsense Knowledge, Ontology and Ordinary Language

**Walid S. Saba**

American Institutes for Research,
1000 Thomas Jefferson Street, NW, Washington, DC 20007 USA
E-mail: wsaba@air.org

**Abstract:** Over two decades ago a "quite revolution" overwhelmingly replaced knowledge-based approaches in natural language processing (NLP) by quantitative (e.g., statistical, corpus-based, machine learning) methods. Although it is our firm belief that purely quantitative approaches cannot be the only paradigm for NLP, dissatisfaction with purely engineering approaches to the construction of large knowledge bases for NLP are somewhat justified. In this paper we hope to demonstrate that both trends are partly misguided and that the time has come to enrich logical semantics with an ontological structure that reflects our commonsense view of the world and the way we talk about in ordinary language. In this paper it will be demonstrated that assuming such an ontological structure a number of challenges in the semantics of natural language (e.g., metonymy, intensionality, copredication, nominal compounds, etc.) can be properly and uniformly addressed.

**Biographical Notes:** W. Saba received his PhD in Computer Science from Carleton University in 1999. He is currently a Principal Software Engineer at the American Institutes for Research in Washington, DC. Prior to this he was in academia where he taught computer science at the University of Windsor and the American University of Beirut (AUB). For over 9 years he was also a consulting software engineer where worked at such places as AT&T Bell Labs, MetLife and Cognos, Inc. His research interests are in natural language processing, ontology, the representation of and reasoning with commonsense knowledge, and intelligent e-commerce agents.

## 1   INTRODUCTION

Over two decades ago a "quite revolution", as Charniak (1995) once called it, overwhelmingly replaced knowledge-based approaches in natural language processing (NLP) by quantitative (e.g., statistical, corpus-based, machine learning) methods. In recent years, however, the terms *ontology*, *semantic web* and *semantic computing* have been in vogue, and regardless of how these terms are being used (or mis-used) we believe that this 'semantic counter revolution' is a positive trend since corpus-based approaches to NLP, while useful in some language processing tasks – see (Ng and Zelle, 1997) for a good review – cannot account for compositionality and productivity in natural language, not to mention the complex inferential patterns that occur in ordinary language use. The inferences we have in mind here can be illustrated by the following example:

(1)   *Pass that car will you.*
        a. *He is really annoying me.*
        b. *They are really annoying me.*

Clearly, speakers of ordinary language can easily infer that 'he' in (1a) refers to the person *driving* [that] car, while 'they' in (1b) is a reference to the people *riding* [that] car. Such inferences, we believe, cannot theoretically be learned (how many such examples will be needed?), and are thus beyond the capabilities of any quantitative approach. On the other hand, and although it is our firm belief that purely quantitative approaches cannot be the only paradigm for NLP, dissatisfaction with purely engineering approaches to the construction of large knowledge bases for NLP (e.g., Lenat and Ghua, 1990) are somewhat justified. While language 'understanding' is for the most part a commonsense 'reasoning' process at the pragmatic level, as example (1) illustrates, the knowledge structures that an NLP system must utilize should have sound linguistic and ontological underpinnings and must be formalized if we ever hope to build scalable systems (or as John McCarthy once said, if we ever hope to build systems that we can actually understand!). Thus, and as we have argued elsewhere (Saba, 2007), we believe that both trends are partly misguided and that the time has come to enrich logical semantics with an ontological structure that reflects our commonsense view of the world and the way we talk about in ordinary language. Specifically, we argue that very little progress within logical semantics have been made in the past several years due to the fact that these systems are, for the most part, mere symbol manipulation systems that are devoid of any content. In particular, in such systems where there is hardly any link between semantics and our commonsense view of the world, it is quite difficult to envision how one can "uncover" the considerable amount of content that is clearly implicit, but almost never explicitly stated in our everyday discourse. For example, consider the following:

(2)   a.   *Simon is a rock.*
      b.   *The ham sandwich wants a beer.*
      c.   *Sheba is articulate.*
      d.   *Jon bought a brick house.*
      e.   *Carlos likes to play bridge.*
      f.   *Jon enjoyed the book.*
      g.   *Jon visited a house on every street.*

Although they tend to use the least number of words to convey a particular thought (perhaps for computational effectiveness, as Givon (1984) once suggested), speakers of ordinary language clearly understand the sentences in (2) as follows:

(3)   a.   *Simon is* [as solid as] *a rock.*
      b.   *The* [person eating the] *ham sandwich wants a beer.*
      c.   *Sheba is* [an] *articulate* [person].
      d.   *Jon bought a brick* [-made] *house.*
      e.   *Carlos likes to play* [the game] *bridge.*
      f.   *Jon enjoyed* [reading/writing] *the book.*
      g.   *Jon visited a* [different] *house on every street.*

Clearly, any compositional semantics must somehow account for this [missing text], as such sentences are quite common and are not at all exotic, farfetched, or contrived. Linguists and semanticists have usually dealt with such sentences by investigating various phenomena such as *metaphor* (3a); *metonymy* (3b); *textual entailment* (3c); *nominal compounds* (3d); *lexical ambiguity* (3e), *co-predication* (3f); and *quantifier scope ambiguity* (3g), to name a few. However, and although they seem to have a common denominator, it is somewhat surprising that in looking at the literature one finds that these phenomena have been studied quite independently; to the point where there is very little, if any, that seems to be common between the various proposals that are often suggested. In our opinion this state of affairs is very problematic, as the prospect of a distinct paradigm for every single phenomenon in natural language cannot be realistically contemplated. Moreover, and as we hope to demonstrate in this paper, we believe that there is indeed a common symptom underlying these (and other) challenging problems in the semantics of natural language.

Before we make our case, let us at this very early juncture suggest this informal explanation for the missing text in (2): SOLID is (one of) the most salient features of a Rock (2a); people, and not a sandwich, have 'wants' and EAT is the most salient relation that holds between a Human and a Sandwich (2b)[1]; Human is the type of object of which ARTICULATE is the most salient property (2c); MADE-OF is the most salient relation between an Artifact (and consequently a House) and a substance (Brick) (2d); PLAY is the most salient relation that holds between a Human and a Game, and not some structure (and, bridge is a game); and, finally, in the (possible) world that we live in, a House cannot be located on more than one Street. The point of this informal explanation is to suggest that the problem underlying most challenges in the semantics of natural language seems to lie in semantic formalisms that employ logics that are mere abstract symbol manipulation systems; systems that are devoid of any ontological content. What we suggest, instead, is a compositional semantics that is grounded in commonsense metaphysics, a semantics that views "logic as a language"; that is, a logic that has content, and ontological content, in particular, as has been recently and quite convincingly advocated by Cocchiarella (2001).

In the rest of the paper we will first propose a semantics that is grounded in a strongly-typed ontology that reflects our commonsense view of reality and the way we talk about it in ordinary language; subsequently, we will formalize the notion of 'salient property' and 'salient relation' and suggest how a strongly-typed compositional system can possibly utilize such information to explain some complex phenomena in natural language.

## 2   A TYPE SYSTEM FOR ORDINARY LANGUAGE

The utility of enriching the ontology of logic by introducing variables and quantification is well-known. For example, $(p \wedge q) \supset r$ is not even a valid statement in propositional logic, when $p = all\ humans\ are\ mortal$, $q = Socrates\ is\ a\ human$ and $r = Socrates\ is\ mortal$. In first-order logic, however, this inference is easily produced, by exploiting one important aspect of variables, namely, their *scope*. However, and as will shortly be demonstrated, copredication, metonymy and various other problems that are relegated to intensionality in natural language are due the fact that another important aspect of a variable, namely its type, has not been exploited. In particular, much like scope connects various predicates within a formula, when a variable has more than one type in a single scope, type unification is the process by which one can discover implicit relationships that are not explicitly stated, but are in fact implicit in the type hierarchy. To begin with, therefore, we shall first introduce a type system that is assumed in the rest of the paper.

### 2.1   The Tree of Language

In *Types and Ontology* Fred Sommers (1963) suggested several years ago that there is a strongly typed ontology that seems to be implicit in all that we say in ordinary spoken language, where two objects $x$ and $y$ are considered to be of the same type iff the set of monadic predicates that are significantly (that is, truly or falsely but not absurdly) predicable of $x$ is equivalent to the set of predicates that are significantly predicable of $y$. Thus, while they make a references to four distinct classes (sets of objects), for an ontologist interested in the relationship between ontology and natural language, the noun phrases in (4) are ultimately referring to two types only, namely Cat and Number:

(4)  a. *an old cat*
     b. *a black cat*
     c. *an even number*
     d. *a prime number*

---

[1]  In addition to EAT, a Human can of course also BUY, SELL, MAKE, PREPARE, WATCH, or HOLD, etc. a Sandwich. Why EAT might be a more salient relation between a Person and a Sandwich is a question we shall pay considerable attention to below.

In other words, whether we make a reference to an *old cat* or to a *black cat*, in both instances we are ultimately speaking of objects that are of the same type; and this, according to Sommers, is a reflection of the fact that the set of monadic predicates in our natural language that are significantly predicable of old cats is exactly the same set that is significantly predicable of black cats. Let us say $sp(\mathsf{t},s)$ is true if $s$ is the set of predicates that are significantly predicable of some type $\mathsf{t}$, and let $\mathbf{T}$ represent the set of all types in our ontology, then

(5)  a. $\mathsf{t} \in \mathbf{T} \equiv (\exists s)[sp(\mathsf{t},s) \wedge (s \neq \phi)]$
     b. $\mathsf{s} \sqsubseteq \mathsf{t} \equiv (\exists s_1, s_2)[sp(\mathsf{s},s_1) \wedge sp(\mathsf{t},s_2) \wedge (s_1 \subseteq s_2)]$
     c. $\mathsf{s} = \mathsf{t} \equiv (\exists s_1, s_2)[sp(\mathsf{s},s_1) \wedge sp(\mathsf{t},s_2) \wedge (s_1 = s_2)]$

That is, to be a type (in the ontology) is to have a non-empty set of predicates that are significantly predicable (5a) [2]; and a type $\mathsf{s}$ is a subtype of $\mathsf{t}$ iff the set of predicates that are significantly predicable of $\mathsf{s}$ is a subset of the set of predicates that are significantly predicable of $\mathsf{t}$ (5b); consequently, the identity of a concept (and thus concept similarity) is well-defined as given by (5c). Note here that according to (5a), abstract objects such as *events*, *states*, *properties*, *activities*, *processes*, etc. are also part of our ontology since the set of predicates that is significantly predicable of any such object is not empty. For example, one can always speak of an *imminent* event, or an event that was *cancelled*, etc., that is $sp(\mathsf{Event}, \{\mathsf{IMMINENT}, \mathsf{CANCELLED}, etc.\})$. In addition to events, abstract objects such as states and processes, etc. can also be predicated; for example, one can always say *idle* of a some state, and one always speak of *starting* and *terminating* a process, etc.

In our representation, therefore, concepts belong to two quite distinct categories: (*i*) ontological concepts, such as Animal, Substance, Entity, Artefact, Event, State, etc., which are assumed to exist in a subsumption hierarchy, and where the fact that an object of type Human is (ultimately) an object of type Entity is expressed as $\mathsf{Human} \sqsubseteq \mathsf{Entity}$ ; and (*ii*) logical concepts, which are the properties (that can be said) of and the relations (that can hold) between ontological concepts. To illustrate the difference (and the relation) between the two, consider the following:

---

[2] Interestingly, (5a) seems to be related to what Fodor (1998) meant by "to be a concept is to be locked to a property"; in that it seems that a genuine concept (or a Sommers' type) is one that 'owns' at least one word/predicate in the language.

(6)  $r_1 : \mathrm{OLD}(x :: \mathsf{Entity})$
     $r_2 : \mathrm{HEAVY}(x :: \mathsf{Physical})$
     $r_3 : \mathrm{HUNGRY}(x :: \mathsf{Living})$
     $r_4 : \mathrm{ARTICULATE}(x :: \mathsf{Human})$
     $r_5 : \mathrm{MAKE}(x :: \mathsf{Human}, y :: \mathsf{Artifact})$
     $r_6 : \mathrm{MANUFACTURE}(x :: \mathsf{Human}, y :: \mathsf{Instrument})$
     $r_7 : \mathrm{RIDE}(x :: \mathsf{Human}, y :: \mathsf{Vehicle})$
     $r_8 : \mathrm{DRIVE}(x :: \mathsf{Human}, y :: \mathsf{Car})$

The predicates in (6) are supposed to reflect the fact that in ordinary spoken we language we can say $\mathrm{OLD}$ of any Entity; that we say $\mathrm{HEAVY}$ of objects that are of type Physical; that $\mathrm{HUNGRY}$ is said of objects that are of type Living; that $\mathrm{AR}$-$\mathrm{TICULATE}$ is said of objects that must be of type Human; that $\mathrm{MAKE}$ is a relation that can hold between a Human and an Artefact; that $\mathrm{MANUFACTURE}$ is a relation that can hold between a Human and an Instrument, etc. Note that the type assignments in (6) implicitly define a type hierarchy as that shown in figure 1 below. Consequently, and although not explicitly stated in (6), in ordinary spoken language one can always attribute the property $\mathrm{HEAVY}$ to an object of type Car since $\mathsf{Car} \sqsubseteq \mathsf{Vehicle} \sqsubseteq \cdots \sqsubseteq \mathsf{Physical}$ .

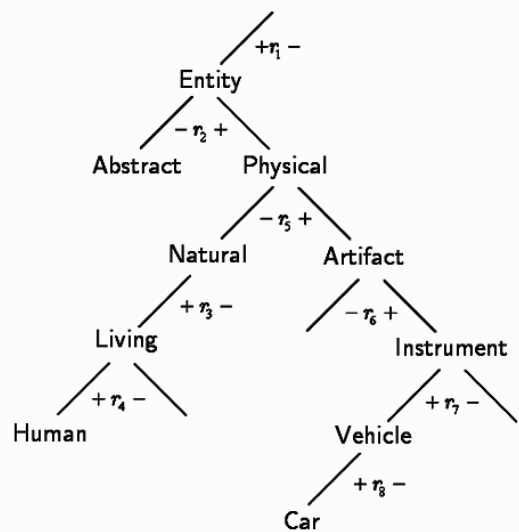

**Figure 1** The type hierarchy implied by (6)

In addition to logical and ontological concepts, there are also proper nouns, which are the names of objects; objects that could be of any type. A proper noun, such as *sheba*, is interpreted as

(7)  $[\![sheba]\!]$
     $\Rightarrow \lambda P[(\exists^1 x)(\mathrm{NOO}(x :: \mathsf{Thing}, 'sheba') \wedge P(x :: \mathsf{t}))]$

where $\mathrm{NOO}(x :: \mathsf{Thing}, s)$ is true of some individual object $x$ (which could be any Thing), and $s$ if (the label) $s$ is the name of $x$, and $\mathsf{t}$ is presumably the type of objects that $P$ applies to (to simplify notation, however, we will often write (7) as $[\![sheba]\!] \Rightarrow \lambda P[(\exists^1 sheba :: \mathsf{Thing})(P(sheba :: \mathsf{t}))]$ ). Consider now the following, where $\text{TEACHER}(x :: \text{Human})$, that is, where $\text{TEACHER}$ is assumed to be a property that is ordinarily said of objects that must be of type Human, and where $\mathbf{BE}(x, y)$ is true when $x$ and $y$ are the same objects[3]:

(8)  $\llbracket sheba \ is \ a \ teacher \rrbracket$
$\Rightarrow (\exists^1 sheba :: \text{Thing})(\exists x)$
$\quad (\text{TEACHER}(x :: \text{Human}) \wedge \mathbf{BE}(sheba, x))$

This states that there is a unique object named *sheba* (which is an object that could be any Thing), and some $x$ such that $x$ is a $\text{TEACHER}$ (and thus must be an object of type Human), and such that *sheba* is that $x$. Since $\mathbf{BE}(sheba, x)$, we can replace $y$ by the constant *sheba* obtaining the following:

(9)  $\llbracket sheba \ is \ a \ teacher \rrbracket$
$\Rightarrow (\exists^1 sheba :: \text{Thing})(\exists x)$
$\quad (\text{TEACHER}(x :: \text{Human}) \wedge \mathbf{BE}(sheba, x))$
$\Rightarrow (\exists^1 sheba :: \text{Thing})(\text{TEACHER}(sheba :: \text{Human}))$

Note now that *sheba* is associated with more than one type in a single scope. In these situations a type unification must occur, where a type unification $(\mathsf{s} \bullet \mathsf{t})$ between two types $\mathsf{s}$ and $\mathsf{t}$ and where $Q \in \{\exists, \forall\}$, is defined (for now) as follows

(10)

$(Qx :: (\mathsf{s} \bullet \mathsf{t}))(P(x))$
$\equiv \begin{cases} (Qx :: \mathsf{s})(P(x)), & if \ (\mathsf{s} \sqsubseteq \mathsf{t}) \\ (Qx :: \mathsf{t})(P(x)), & if \ (\mathsf{t} \sqsubseteq \mathsf{s}) \\ (Qx :: \mathsf{s})(Qy :: \mathsf{t})(\mathbf{R}(x, y) \wedge P(y)), & if \ (\exists \mathbf{R})(\mathbf{R} = msr(\mathsf{s}, \mathsf{t})) \\ \bot, & otherwise \end{cases}$

where $\mathbf{R}$ is some salient relation that might exist between objects of type $\mathsf{s}$ and objects of type $\mathsf{t}$. That is, in situations where there is no subsumption relation between $\mathsf{s}$ and $\mathsf{t}$ the type unification results in keeping the variables of both types and in introducing some salient relation between them (we shall discuss these situations below).

Going to back to (9), the type unification in this case is actually quite simple, since $(\text{Human} \sqsubseteq \text{Thing})$:

(11)  $\llbracket sheba \ is \ a \ teacher \rrbracket$
$\Rightarrow (\exists^1 sheba :: \text{Thing})(\exists x)(\text{TEACHER}(sheba :: \text{Human}))$
$\Rightarrow (\exists^1 sheba :: (\text{Thing} \bullet \text{Human}))(\text{TEACHER}(sheba))$
$\Rightarrow (\exists^1 sheba :: \text{Human})(\text{TEACHER}(sheba))$

In the final analysis, therefore, *sheba is a teacher* is interpreted as follows: there is a unique object named *sheba*, an object that must be of type Human, such that *sheba* is a $\text{TEACHER}$. Note here the clear distinction between ontological concepts (such as Human), which Cocchiarella (2001) calls first-intension concepts, and logical (or second-intension) concepts, such as $\text{TEACHER}(x)$. That is, what ontologically exist are objects of type Human, not teachers, and

---

[3]  We are using the fact that, when $a$ is a constant and $P$ is a predicate, $Pa \equiv \exists x [Px \wedge (x = a)]$ (see Gaskin, 1995).

$\text{TEACHER}$ is a mere property that we have come to use to talk of objects of type Human[4]. In other words, while the property of being a $\text{TEACHER}$ that $x$ may exhibit is accidental (as well as temporal, cultural-dependent, etc.), the fact that some $x$ is an object of type Human (and thus an Animal, etc.) is not. Moreover, a logical concept such as $\text{TEACHER}$ is assumed to be defined by virtue of some logical expression such as $(\forall x :: \text{Human})(\text{TEACHER}(x) \equiv_{df} \varphi)$, where the exact nature of $\varphi$ might very well be susceptible to temporal, cultural, and other contextual factors, depending on what, at a certain point in time, a certain community considers a $\text{TEACHER}$ to be. Specifically, the logical concept $\text{TEACHER}$ must be defined by some expression such as

$(\forall x :: \text{Human})(\text{TEACHER}(x)$
$\equiv_{df} (\exists a :: \text{Activity})(\text{TEACHING}(a) \wedge \text{AGENT}(a, x)))$

That is, any $x$, which must be an object of type Human, is a $\text{TEACHER}$ iff $x$ is the agent of some Activity $a$, where $a$ is a $\text{TEACHING}$ activity. It is certainly not for convenience, elegance or mere ontological indulgence that a logical concept such as $\text{TEACHER}$ must be defined in terms of more basic ontological categories (such as an Activity) as can be illustrated by the following example:

(12)  $\llbracket sheba \ is \ a \ superb \ teacher \rrbracket$
$\Rightarrow (\exists^1 sheba :: \text{Thing})(\text{SUPERB}(sheba :: \text{Human})$
$\quad \wedge \text{TEACHER}(sheba :: \text{Human}))$

Note that in (12), it is *sheba*, and not her teaching that is erroneously considered to be *superb*. This is problematic on two grounds: first, while $\text{SUPERB}$ is a property that could apply to objects of type Human (such as *sheba*), the logical form in (12) must have a reference to an object of type Activity, as $\text{SUPERB}$ is a property that could also be said of *sheba*'s teaching activity. This point is more acutely made when *superb* is replaced by adjectives such as *certified*, *lousy*, etc., where the corresponding properties do not even apply to *sheba*, but are clearly modifying *sheba*'s teaching activity (that it is $\text{CERTIFIED}$, or $\text{LOUSY}$, etc.) We shall discuss this issue in some detail below. Before we proceed, however, we need to extend the notion of type unification slightly.

## 2.2  More on Type Unification

It should be clear by now that our ontology, as defined thus far, assumes a Platonic universe which admits the existence of anything that can be talked about in ordinary language. Thus, and as also argued by Cocchiarella (1996), besides abstract objects, reference in ordinary language can be made to objects that might have or could have existed, as well as to objects that might exist sometime in the future. In general, therefore, a reference to an object can be[5]

---

[4]  Not recognizing the difference between logical (e.g., $\text{TEACHER}$) and ontological concepts (e.g., Human) is perhaps the reason why ontologies in most AI systems are rampant with multiple inheritance.
[5]  We can use $\Diamond a$ to state that an object is possibly abstract, instead of $\neg \Box c$, which is intended to state that the object is not necessarily concrete (or that it does not necessarily actually exist).

- a reference to a type (in the ontology): $(\exists X :: \mathsf{t})(P(X))$;
- a reference to an object of a certain type, an object that must have a concrete existence: $(\exists X :: \mathsf{t})(P(X^{\square c}))$; or
- a reference to an object of a certain type, an object that need not actually exist: $(\exists X :: \mathsf{t})(P(X^{\neg\square c}))$.

Accordingly, and as suggested by Hobbs (1985), the above necessitates that a distinction be made in our logical form between mere *being* and concrete (or actual) existence. To do this we introduce a predicate $Exist(x)$ which is true when some object $x$ has a concrete (or actual) existence, and where a reference to an object of some type is initially assumed to be imply mere being, while actual (or concrete) existence is only inferred from the context. The relationship between mere being and concrete existence can be defined as follows:

(13)   a. $(\exists X :: \mathsf{t})(P(X))$

     b. $(\exists X :: \mathsf{t})(P(X^{\square c}))$
$\equiv (\exists X :: \mathsf{t})(\exists x)(Inst(x, X) \wedge Exist(x) \wedge P(x))$

     c. $(\exists X :: \mathsf{t})(P(X^{\neg\square c}))$
$\equiv (\exists X :: \mathsf{t})(\forall x)(Inst(x, X) \wedge Exist(x) \supset P(x))$

In (13a) we are simply stating that some property $P$ is true of some object $X$ of type t. Thus, while, ontologically, there are objects of type t that we can speak about, nothing in (13a) entails the actual (or concrete) existence of any such objects. In (13b) we are stating that the property $P$ is true of an object $X$ of type t, an object that must have a concrete (or actual) existence (and in particular at least the instance $x$); which is equivalent to saying that there is some object $x$ which is an instance of some abstract object $X$, where $x$ actually exists, and where $P$ is true of $x$. Finally, (13c) states that whenever some $x$, which is an instance of some abstract object $X$ of type t exists, then the property $P$ is true of $x$. Thus, while (13a) make a reference to a *kind* (or a type in the ontology), (13b) and (13c) make a reference to some instance of a specific type, an instance that may or may not actually exist. To simplify notation, therefore, we can write (13b) and (13c) as follows, respectively:

$(\exists X :: \mathsf{t})(P(X^{\square c}))$
$\equiv (\exists X :: \mathsf{t})(\exists x)(Inst(x, X) \wedge Exist(x) \wedge P(X))$
$\equiv (\exists x :: \mathsf{t})(Exist(x) \wedge P(x))$

$(\exists X :: \mathsf{t})(P(X^{\neg\square c}))$
$\equiv (\exists X :: \mathsf{t})(\forall x)(Inst(x, X) \wedge Exist(x) \supset P(x))$
$\equiv (\forall x :: \mathsf{t})(Exist(x) \supset P(x))$

Furthermore, it should be noted that $x$ in (13b) is assumed to have actual/concrete existence assuming that the property/relation $P$ is actually true of $x$. If the truth of $P(X)$ is just a possibility, then so is the concrete existence of some instance $x$ of $X$. Formally, we have the following:

$(\exists X :: \mathsf{t})(\mathbf{can}(P(X^{\square c}))) \equiv (\exists X :: \mathsf{t})(P^{\mathbf{can}}(X^{\neg\square c}))$

Finally, and since different relations and properties have different existence assumptions, the existence assumptions implied by a compound expression is determined by type unification, which is defined as follows, and where the basic type unification $(\mathsf{s} \bullet \mathsf{t})$ is that defined in (10):

$(x :: (\mathsf{s} \bullet \mathsf{t}^{\square c})) = (x :: (\mathsf{s} \bullet \mathsf{t})^{\square c})$
$(x :: (\mathsf{s} \bullet \mathsf{t}^{\neg\square c})) = (x :: (\mathsf{s} \bullet \mathsf{t})^{\neg\square c})$
$(x :: (\mathsf{s}^{\square c} \bullet \mathsf{t}^{\neg\square c})) = (x :: (\mathsf{s} \bullet \mathsf{t})^{\square c})$

As a first example consider the following (where temporal and modal auxiliaries are represented as superscripts on the predicates):

(14)   $[\![\, jon\ needs\ a\ computer \,]\!]$
$\Rightarrow (\exists^1 jon :: \mathsf{Human})(\exists X :: \mathsf{Computer})$
$\quad (\mathrm{NEED}^{\mathbf{does}}(jon, X :: \mathsf{Thing}))$

In (14) we are stating that some unique object named *jon*, which is of type Human does NEED something we call Computer. On the other hand, consider now the interpretation of '*jon fixed a computer*':

(15)   $[\![\, jon\ fixed\ a\ computer \,]\!]$
$\Rightarrow (\exists^1 jon :: \mathsf{Human})(\exists X :: \mathsf{Computer})$
$\quad (\mathrm{FIX}^{\mathbf{did}}(jon, X :: \mathsf{Thing}^{\square c}))$
$\Rightarrow (\exists^1 jon :: \mathsf{Human})(\exists X :: (\mathsf{Computer} \bullet \mathsf{Thing}^{\square c}))$
$\quad (\mathrm{FIX}^{\mathbf{did}}(jon, X))$
$\Rightarrow (\exists^1 jon :: \mathsf{Human})(\exists X :: \mathsf{Computer}^{\square c})$
$\quad (\mathrm{FIX}^{\mathbf{did}}(jon, X))$
$\Rightarrow (\exists^1 jon :: \mathsf{Human})(\exists X :: \mathsf{Computer})$
$\quad (\exists x)(Inst(x, X) \wedge Exist(x) \wedge \mathrm{FIX}^{\mathbf{did}}(jon, X))$
$\Rightarrow (\exists^1 jon :: \mathsf{Human})(\exists x :: \mathsf{Computer})$
$\quad (Exist(x) \wedge \mathrm{FIX}^{\mathbf{did}}(jon, x))$

That is, '*jon fixed a computer*' is interpreted as follows: there is a unique object named *jon*, which is an object of type Human, and some $x$ of type Computer (an $x$ that actually exists) such that *jon* did FIX $x$. However, consider now the following:

(16)   $[\![\, jon\ can\ fix\ a\ computer \,]\!]$
$\Rightarrow (\exists^1 jon :: \mathsf{Human})(\exists X :: \mathsf{Computer})$
$\quad (\mathrm{FIX}^{\mathbf{can}}(jon, X :: \mathsf{Thing}^{\neg\square c}))$
$\Rightarrow (\exists^1 jon :: \mathsf{Human})(\exists X :: (\mathsf{Computer} \bullet \mathsf{Thing}^{\neg\square c}))$
$\quad (\mathrm{FIX}^{\mathbf{can}}(jon, X))$
$\Rightarrow (\exists^1 jon :: \mathsf{Human})(\exists X :: \mathsf{Computer}^{\neg\square c})$
$\quad (\mathrm{FIX}^{\mathbf{can}}(jon, X))$
$\Rightarrow (\exists^1 jon :: \mathsf{Human})(\exists X :: \mathsf{Computer})$
$\quad (\forall x)(Inst(x, X) \wedge Exist(x) \supset \mathrm{FIX}^{\mathbf{can}}(jon, X))$
$\Rightarrow (\exists^1 jon :: \mathsf{Human})(\exists x :: \mathsf{Computer})$
$\quad (\forall x)(Exist(x) \supset \mathrm{FIX}^{\mathbf{can}}(jon, x))$

Essentially, therefore, '*jon can fix a computer*' is stating that whenever an object $x$ of type Computer exists, then *jon* can fix $x$; or, equivalently, that '*jon can fix* any *computer*'.

Finally, consider the following, where it is assumed that our ontology reflects the commonsense fact that we can always speak of an Animal climbing some Physical object:

$[\![\, a\ snake\ can\ climb\ a\ tree \,]\!]$

$$\Rightarrow (\exists X :: \text{Snake})(\exists Y :: \text{Tree})$$
$$(\text{CLIMB}^{\mathbf{can}}(X :: \text{Animal}^{\sqsubset C}, Y :: \text{Physical}^{\sqsubset C}))$$
$$\Rightarrow (\exists X :: (\text{Snake} \bullet \text{Animal}^{\sqsubset C}))(\exists Y :: (\text{Tree} \bullet \text{Physical}^{\sqsubset C}))$$
$$(\text{CLIMB}^{\mathbf{can}}(X, Y))$$
$$\Rightarrow (\exists X :: \text{Snake}^{\sqsubset C})(\exists Y :: \text{Tree}^{\sqsubset C})(\text{CLIMB}^{\mathbf{can}}(X, Y))$$
$$\Rightarrow (\exists X :: \text{Snake})(\exists Y :: \text{Tree})$$
$$(\forall x)(\forall y)(Inst(x, X) \wedge Exist(x) \wedge Inst(y, Y)$$
$$\wedge \, Exist(y) \supset \text{CLIMB}^{\mathbf{can}}(x, y))$$
$$\Rightarrow (\forall x :: \text{Snake})(\forall y :: \text{Tree})$$
$$(Exist(x) \wedge Exist(y) \supset \text{CLIMB}^{\mathbf{can}}(x, y))$$

That is, '*a snake can climb a tree*' is essentially interpreted as *any snake* (if it exists) *can climb any tree* (if it exists).

With this background, we now proceed to tackle some interesting problems in the semantics of natural language.

## 3   SEMANTICS WITH ONTOLOGICAL CONTENT

In this section we discuss several problems in the semantic of natural language and demonstrate the utility of a semantics embedded in a strongly-typed ontology that reflects our commonsense view of reality and the way we take about it in ordinary language.

### 3.1   Types, Polymorphism and Nominal Modification

We first demonstrate the role type unification and polymorphism plays in nominal modification. Consider the sentence in (1) which could be uttered by someone who believes that: (*i*) Olga is a dancer and a beautiful person; or (*ii*) Olga is beautiful as a dancer (i.e., Olga is a dancer and she dances beautifully).

(17)   *Olga is a beautiful dancer*

As suggested by Larson (1998), there are two possible routes to explaining this ambiguity: one could assume that a noun such as 'dancer' is a simple one place predicate of type $\langle e, t \rangle$ and 'blame' this ambiguity on the adjective; alternatively, one could assume that the adjective is a simple one place predicate and blame the ambiguity on some sort of complexity in the structure of the head noun (Larson calls these alternatives *A*-analysis and *N*-analysis, respectively).

In an *A*-analysis, an approach advocated by Siegel (1976), adjectives are assumed to belong to two classes, termed predicative and attributive, where predicative adjectives (e.g., *red*, *small*, etc.) are taken to be simple functions from entities to truth-values, and are thus extensional and intersective: $[\![ Adj \, Noun ]\!] = [\![ Adj ]\!] \cap [\![ Noun ]\!]$. Attributive adjectives (e.g., *former*, *previous*, *rightful*, etc.), on the other hand, are functions from common noun denotations to common noun denotations – i.e., they are predicate modifiers of type $\langle \langle e, t \rangle, \langle e, t \rangle \rangle$, and are thus intensional and non-intersective (but subsective: $[\![ Adj \, Noun ]\!] \subseteq [\![ Noun ]\!]$). On this view, the ambiguity in (17) is explained by posting two distinct lexemes ($beautiful_1$ and $beautiful_2$) for the adjective *beautiful*, one of which is an attributive while the other is a predicative adjective. In keeping with Montague's

(1970) edict that similar syntactic categories must have the same semantic type, for this proposal to work, all adjectives are initially assigned the type $\langle \langle e, t \rangle, \langle e, t \rangle \rangle$ where intersective adjectives are considered to be subtypes obtained by triggering an appropriate meaning postulate. For example, assuming the lexeme $beautiful_1$ is marked (for example by a lexical feature such as +INTERSECTIVE), then the meaning postulate $\exists P \forall Q \forall x [\text{BEAUTIFUL}(Q)(x) \leftrightarrow P(x) \wedge Q(x)]$ does yield an intersective meaning when $P$ is $beautiful_1$; and where a phrase such as `a beautiful dancer' is interpreted as follows[6]:

$$[\![ a \, beautiful_1 \, dancer ]\!]$$
$$\Rightarrow \lambda P[(\exists x)(\text{DANCER}(x) \wedge \text{BEAUTIFUL}(x) \wedge P(x))]$$
$$[\![ a \, beautiful_2 \, dancer ]\!]$$
$$\Rightarrow \lambda P[(\exists x)(\text{BEAUTIFUL}(\hat{\,}\text{DANCER}(x)) \wedge P(x))]$$

While it does explain the ambiguity in (17), several reservations have been raised regarding this proposal. As Larson (1995; 1998) notes, this approach entails considerable duplication in the lexicon as this means that there are 'doublets' for all adjectives that can be ambiguous between an intersective and a non-intersective meaning. Another objection, raised by McNally and Boleda (2004), is that in an A-analysis there are no obvious ways of determining the context in which a certain adjective can be considered intersective. For example, they suggest that the most natural reading of (18) is the one where *beautiful* is describing Olga's dancing, although it does not modify any noun and is thus wrongly considered intersective by modifying Olga.

(18)   *Look at Olga dance. She is beautiful.*

While valid in other contexts, in our opinion this observation does not necessarily hold in this specific example since the resolution of `she' must ultimately consider all entities in the discourse, including, presumably, the dancing activity that would be introduced by a Davidsonian representation of 'Look at Olga dance' (this issue is discussed further below).

A more promising alternative to the *A*-analysis of the ambiguity in (17) has been proposed by Larson (1995, 1998), who suggests that *beautiful* in (17) is a simple intersective adjective of type $\langle e, t \rangle$ and that the source of the ambiguity is due to a complexity in the structure of the head noun. Specifically, Larson suggests that a deverbal noun such as *dancer* should have the Davidsonian representation $(\forall x)(\text{DANCER}(x) =_{df} (\exists e)(\text{DANCING}(e) \wedge \text{AGENT}(e, x)))$ i.e., any $x$ is a dancer iff $x$ is the agent of some dancing activity (Larson's notation is slightly different). In this analysis, the ambiguity in (1) is attributed to an ambiguity in what *beautiful* is modifying, in that it could be said of Olga or her dancing Activity. That is, (17) is to be interpreted as follows:

$$[\![ Olga \, is \, a \, beautiful \, dancer ]\!]$$
$$\Rightarrow (\exists e)(\text{DANCING}(e) \wedge \text{AGENT}(e, olga)$$
$$\wedge (\text{BEAUTIFUL}(e) \vee \text{BEAUTIFUL}(olga)))$$

---

[6] Note that as an alternative to meaning postulates that specialize intersective adjectives to $\langle e, t \rangle$, one can perform a type-lifting operation from $\langle e, t \rangle$ to $\langle \langle e, t \rangle, \langle e, t \rangle \rangle$ (see Partee, 2007).

In our opinion, Larson's proposal is plausible on several grounds. First, in Larson's N-analysis there is no need for impromptu introduction of a considerable amount of lexical ambiguity. Second, and for reasons that are beyond the ambiguity of beautiful in (17), and as argued in the interpretation of example (12) above, there is ample evidence that the structure of a deverbal noun such as *dancer* must admit a reference to an abstract object, namely a dancing Activity; as, for example, in the resolution of 'that' in (19).

(19)    *Olga is an old dancer.*
        *She has been doing* that *for 30 years.*

Furthermore, and in addition to a plausible explanation of the ambiguity in (17), Larson's proposal seems to provide a plausible explanation for why 'old' in (4a) seems to be ambiguous while the same is not true of 'elderly' in (4b): `old' could be said of Olga or her teaching; while *elderly* is not an adjective that is ordinarily said of objects that are of type activity:

(20)    a. *Olga is an old dancer.*
        b. *Olga is an elderly teacher.*

With all its apparent appeal, however, Larson's proposal is still lacking. For one thing, and it presupposes that some sort of type matching is what ultimately results in rejecting the subsective meaning of *elderly* in (20b), the details of such processes are more involved than Larson's proposal seems to imply. For example, while it explains the ambiguity of *beautiful* in (17), it is not quite clear how an *N-Analysis* can explain why *beautiful* does not seem to admit a subsective meaning in (21).

(21)    *Olga is a beautiful young street dancer.*

In fact, *beautiful* in (21) seems to be modifying Olga for the same reason the sentence in (22a) seems to be more natural than that in (22b).

(22)    a. *Maria is a clever young girl.*
        b. *Maria is a young clever girl.*

The sentences in (22) exemplify what is known in the literature as adjective ordering restrictions (AORs). However, despite numerous studies of AORs (e.g., see Wulff, 2003; Teodorescu, 2006), the slightly differing AORs that have been suggested in the literature have never been formally justified. What we hope to demonstrate below however is that the apparent ambiguity of some adjectives and adjective-ordering restrictions are both related to the nature of the ontological categories that these adjectives apply to in ordinary spoken language. Thus, and while the general assumptions in Larson's (1995; 1998) *N-Analysis* seem to be valid, it will be demonstrated here that nominal modification seem to be more involved than has been suggested thus far. In particular, it seems that attaining a proper semantics for nominal modification requires a much richer type system than currently employed in formal semantics.

First let us begin by showing that the apparent ambiguity of an adjective such as *beautiful* is essentially due to the fact that *beautiful* applies to a very generic type that subsumes many others. Consider the following, where we assume $\text{BEAUTIFUL}(x :: \text{Entity})$ ; that is that BEAUTIFUL can be said of any Entity:

$$[\![Olga\ is\ a\ beautiful\ dancer]\!]$$
$$\Rightarrow (\exists^1 Olga :: \text{Human})(\exists a :: \text{Activity})$$
$$(\text{DANCING}(a) \wedge \text{AGENT}(a, Olga :: \text{Human}) \wedge$$
$$(\text{BEAUTIFUL}(a :: \text{Entity}) \vee \text{BEAUTIFUL}(Olga :: \text{Entity}))$$

Note now that, in a single scope, $a$ is considered to be an object of type Activity as well as an object of type Entity, while $Olga$ is considered to be a Human and an Entity. This, as discussed above, requires a pair of type unifications, $(\text{Human} \sqsubseteq \text{Entity})$ and $(\text{Activity} \sqsubseteq \text{Entity})$ . In this case both type unifications succeed, resulting in Human and Activity, respectively:

$$[\![Olga\ is\ a\ beautiful\ dancer]\!]$$
$$\Rightarrow (\exists^1 Olga :: \text{Human})(\exists a :: \text{Activity})$$
$$(\text{DANCING}(a) \wedge \text{AGENT}(a, Olga)$$
$$\wedge (\text{BEAUTIFUL}(a) \vee \text{BEAUTIFUL}(Olga)))$$

In the final analysis, therefore, 'Olga is a beautiful dancer' is interpreted as: Olga is the agent of some dancing Activity, and either Olga is BEAUTIFUL or her DANCING (or, of course, both). However, consider now the following, where ELDERLY is assumed to be a property that applies to objects that must be of type Human:

$$[\![Olga\ is\ an\ elderly\ teacher]\!]$$
$$\Rightarrow (\exists^1 Olga :: \text{Human})(\exists a :: \text{Activity})$$
$$(\text{TEACHING}(a) \wedge \text{AGENT}(a, Olga :: \text{Human}) \wedge$$
$$(\text{ELDERLY}(a :: \text{Human}) \vee \text{ELDERLY}(Olga :: \text{Human})))$$

Note now that the type unification concerning Olga is trivial, while the type unification concerning $a$ will fail since $(\text{Activity} \bullet \text{Human}) = \bot$, thus resulting in the following:

$$[\![Olga\ is\ an\ elderly\ teacher]\!]$$
$$\Rightarrow (\exists^1 Olga :: \text{Human})(\exists a :: \text{Activity})$$
$$(\text{TEACHING}(a) \wedge \text{AGENT}(a, Olga :: \text{Human})$$
$$\wedge (\text{ELDERLY}(a :: (\text{Human} \bullet \text{Activity}))$$
$$\vee \text{ELDERLY}(Olga :: \text{Human}))$$
$$\Rightarrow (\exists^1 Olga :: \text{Human})(\exists a :: \text{Activity})(\text{TEACHING}(a)$$
$$\wedge \text{AGENT}(a, Olga) \wedge (\bot \vee \text{ELDERLY}(Olga))$$
$$\Rightarrow (\exists^1 Olga :: \text{Human})(\exists a :: \text{Activity})$$
$$(\text{TEACHING}(a) \wedge \text{AGENT}(a, Olga) \wedge \text{ELDERLY}(Olga))$$

Thus, in the final analysis, '*Olga is an elderly teacher*' is interpreted as follows: there is a unique object named *Olga*, an object that must be of type Human, and an object $a$ of type Activity, such that $a$ is a teaching activity, Olga is the agent of the activity, and such that elderly is true of *Olga*.

## 3.2  Adjective Ordering Restrictions

Assuming $\text{BEAUTIFUL}(x :: \text{Entity})$ - i.e., that *beautiful* is a property that can be said of objects of type Entity, then it is a property that can be said of a Cat, a Person, a City, a Movie, a Dance, an Island, etc. Therefore, BEAUTIFUL can be thought of as a polymorphic function that applies to objects at several levels and where the semantics of this function depend on the type of the object, as illustrated in figure 2 below[7]. Thus, and although BEAUTIFUL applies to objects of type Entity, in saying 'a beautiful car', for example, the meaning of *beautiful* that is accessed is that defined in the type Physical (which could in principal be inherited from a supertype). Moreover, and as is well known in the theory of programming languages, one can always perform type casting upwards, but not downwards (e.g., one can always view a Car as just an Entity, but the converse is not true)[8].

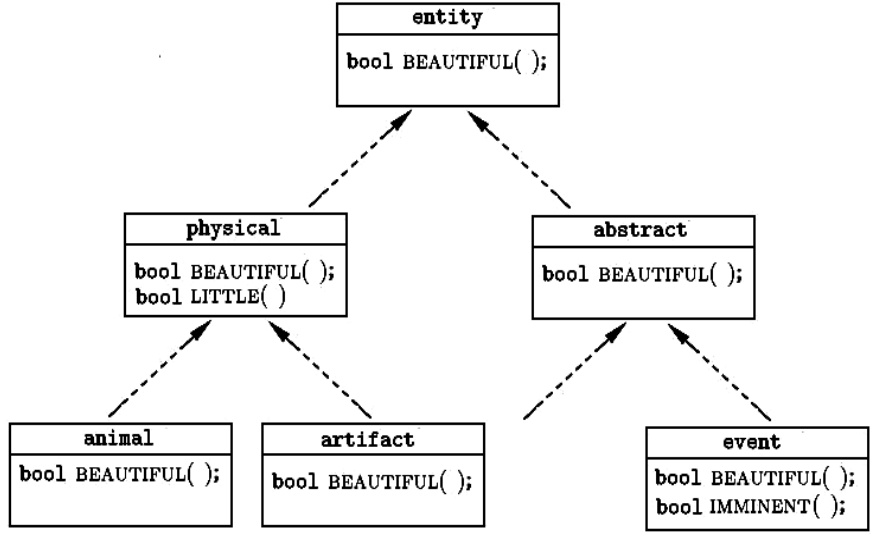

**Figure 2.** Adjectives as polymorphic functions

Thus, and assuming also that RED($x$ :: Physical) ; that is, assuming that RED can be said of Physical objects, then, for example, the type casting that will be required in (23a) is valid, while that in (23b) is not.

(23)  a.  BEAUTIFUL(RED($x$ :: Physical) :: Entity)
      b.  RED(BEAUTIFUL($x$ :: Entity) :: Physical)

This, in fact, is precisely why '*Jon owns a beautiful red car*', for example, is more natural than '*Jon owns a red beautiful car*'. In general, a sequence $A_1(A_2(x :: s) :: t)$ is a valid sequence iff $(s \sqsubseteq t)$. Note that this is different from type unification, in that the unification does succeed in both cases in (11). However, before we perform type unification

the direction of the type casting must be valid. For example, consider the following:

⟦*Olga is a beautiful young dancer*⟧
$\Rightarrow (\exists^1 Olga :: \mathsf{Human})(\exists a :: \mathsf{Activity})$
$\quad (\mathsf{DANCING}(a) \wedge \mathsf{AGENT}(a, Olga) \wedge$
$\quad\quad (\mathsf{BEAUTIFUL}(\mathsf{YOUNG}(a :: \mathsf{Activity}) :: \mathsf{Physical}) :: \mathsf{Entity})$
$\quad\quad \vee \mathsf{BEAUTIFUL}(\mathsf{YOUNG}(Olga :: \mathsf{Human})$
$\quad\quad\quad :: \mathsf{Physical}) :: \mathsf{Entity}))$

Note now that the type casting required (and thus the order of adjectives) is valid since $(\mathsf{Physical} \sqsubseteq \mathsf{Entity})$. This means that we can now perform the required type unifications which would proceed as follows:

$\Rightarrow (\exists^1 Olga :: \mathsf{Human})(\exists a :: \mathsf{Activity})$
$\quad (\mathsf{DANCING}(a) \wedge \mathsf{AGENT}(a, Olga) \wedge$
$\quad\quad (\mathsf{BEAUTIFUL}(\mathsf{YOUNG}(a :: \mathsf{Activity}) :: \mathsf{Physical}) :: \mathsf{Entity})$
$\quad\quad \vee \mathsf{BEAUTIFUL}(\mathsf{YOUNG}(Olga :: \mathsf{Human})$
$\quad\quad\quad :: \mathsf{Physical}) :: \mathsf{Entity}))$

Note now that the type casting required (and thus the order of adjectives) is valid since $(\mathsf{Physical} \sqsubseteq \mathsf{Entity})$. This means that we can now perform the required type unifications which would proceed as follows:

⟦*Olga is a beautiful young dancer*⟧
$\Rightarrow (\exists^1 Olga :: \mathsf{Human})(\exists a :: \mathsf{Activity}) \wedge \mathsf{AGENT}(a, Olga)$
$\quad \wedge (\mathsf{BEAUTIFUL}(\mathsf{YOUNG}(a :: (\mathsf{Activity} \bullet \mathsf{Physical}))$
$\quad\quad \vee \mathsf{BEAUTIFUL}(\mathsf{YOUNG}(Olga :: (\mathsf{Human} \bullet \mathsf{Physical})))$

Since $(\mathsf{Activity} \bullet \mathsf{Physical}) = \perp$, the term involving this type unification is reduced to $\perp$, and $(\perp \vee \beta)$ to $\beta$, hence:

---

[7] It is perhaps worth investigating the relationship between the number of meanings of a certain adjective (say in a resource such as WordNet), and the number of different functions that one would expect to define for the corresponding adjective.

[8] Technically, the reason we can always cast up is that we can always ignore additional information. Casting down, which entails adding information, is however undecidable.

⟦*Olga is a beautiful young dancer*⟧
$\Rightarrow (\exists^1 Olga :: \text{Human})(\exists a :: \text{Activity}) \wedge \text{AGENT}(a, Olga)$
$\wedge (\text{BEAUTIFUL}(\text{YOUNG}(Olga)))$

Note here that since BEAUTIFUL was preceded by YOUNG, it could have not been applicable to an abstract object of type Activity, but was instead reduced to that defined at the level of Physical, and subsequently to that defined at the type Human. A valid question that comes to mind here is how then do we express the thought 'Olga is a young dancer and she dances beautifully'. The answer is that we usually make a statement such as this:

(24)  *Olga is a young and beautiful dancer.*

Note that in this case we are essentially overriding the sequential processing of the adjectives, and thus the adjective-ordering restrictions (or, equivalently, the type-casting rules!) are no more applicable. That is, (24) is essentially equivalent to two sentences that are processed in parallel:

⟦*Olga is a yong and beautiful dancer*⟧
$\equiv$ ⟦*Olga is a young dancer*⟧
$\wedge$ ⟦*Olga is a beautiful dancer*⟧

Note now that 'beautiful' would again have an intersective and a subsective meaning, although 'young' will only apply to Olga due to type constraints.

### 3.3  Intensional Verbs and Coordination

Consider the following sentences and their corresponding translation into standard first-order logic:

(25)  a. ⟦*jon found a unicorn*⟧
        $\Rightarrow (\exists x)(\text{UNICORN}(x) \wedge \text{FIND}(jon, x))$
      b. ⟦*jon sought a unicorn*⟧
        $\Rightarrow (\exists x)(\text{UNICORN}(x) \wedge \text{SEEK}(jon, x))$

Note that $(\exists x)(\text{UNICORN}(x))$ can be inferred in both cases, although it is clear that 'jon sought a unicorn' should not entail the existence of a unicorn. In addressing this problem, Montague (1960) suggested treating seek as an intensional verb that more or less has the meaning of 'tries to find'; i.e. a verb of type $\langle\langle\langle e, t\rangle, t\rangle, \langle e, t\rangle\rangle$, using the tools of a higher-order intensional logic. To handle contexts where there are intensional as well as extensional verbs, mechanisms such as the 'type lifting' operation of Partee and Rooth (1983) were also introduced. The type lifting operation essentially coerces the types into the lowest type, the assumption being that if 'jon sought and found' a unicorn, then a unicorn that was initially sought, but subsequently found, must have concrete existence.

In addition to unnecessary complication of the logical form, we believe the same intuition behind the 'type lifting' operation, which, as also noted by (Kehler *et. al.*, 1995) and Winter (2007), fails in mixed contexts containing more than tow verbs, can be captured without the a priori separation of verbs into intensional and extensional ones, and in particular since most verbs seem to function intensionally and extensionally depending on the context. To illustrate this point further consider the following, where it is assumed that $\text{PAINT}(x :: \text{Human}, y :: \text{Physical})$; that is, it is assumed that the object of paint does not necessarily (although it might) exist:

(26)  ⟦*jon painted a dog*⟧
        $\Rightarrow (\exists^1 jon :: \text{Human})(\exists D :: \text{Dog})$
            $(\text{PAINT}^{\text{did}}(jon :: \text{Human}, D :: \text{Physical}))$
        $\Rightarrow (\exists^1 jon :: \text{Human})(\exists D :: (\text{Dog} \bullet \text{Physical}))$
            $(\text{PAINT}^{\text{did}}(jon, D))$
        $\Rightarrow (\exists^1 jon :: \text{Human})(\exists D :: \text{Dog})(\text{PAINT}^{\text{did}}(jon, D))$

Thus, '*Jon painted a dog*' simply states that some unique object named *jon*, which is an object of type Human painted something we call a Dog. However, let us now assume $\text{OWN}(x :: \text{Human}, y :: \text{Entity}^{\square C})$; that is, if some Human owns some *y* then *y* must actually exist. Consider now all the steps in the interpretation of '*jon painted his dog*':

(27)

⟦*jon painted his dog*⟧
$\Rightarrow (\exists^1 jon :: \text{Human})(\exists D :: \text{Dog})$
    $(\text{OWN}(jon :: \text{Human}, D :: \text{Physical}^{\square C})$
        $\wedge \text{PAINT}(jon :: \text{Human}, D :: \text{Entity}))$
$\Rightarrow (\exists^1 jon :: \text{Human})(\exists D :: \text{Dog})$
    $(\text{OWN}(jon, D :: (\text{Physical}^{\square C} \bullet \text{Entity})) \wedge \text{PAINT}(jon, D))$
$\Rightarrow (\exists^1 jon :: \text{Human})(\exists D :: \text{Dog})$
    $(\text{OWN}(jon, D :: \text{Physical}^{\square C}) \wedge \text{PAINT}(jon, D))$
$\Rightarrow (\exists^1 jon :: \text{Human})(\exists D :: (\text{Dog} \bullet \text{Physical}^{\square C}))$
    $(\text{OWN}(jon, D) \wedge \text{PAINT}(jon, D))$
$\Rightarrow (\exists^1 jon :: \text{Human})(\exists D :: \text{Dog}^{\square C})$
    $(\text{OWN}(jon, D) \wedge \text{PAINT}(jon, D))$

Thus, that while painting something does not entail its existence, owning something does, and the type unification of the conjunction yields the desired result. As given by the rules concerning existence assumptions given in (13) above, the final interpretation should now be proceed as follows:

⟦*jon painted his dog*⟧
$\Rightarrow (\exists^1 jon :: \text{Human})(\exists D :: \text{Dog})$
    $(\exists d)(Inst(d, D) \wedge Exist(d)$
        $\wedge \text{OWN}(jon, d) \wedge \text{PAINT}(jon, d))$
$\Rightarrow (\exists^1 jon :: \text{Human})(\exists D :: \text{Dog})$
    $(Exist(d) \wedge \text{OWN}(jon, d) \wedge \text{PAINT}(jon, d))$

That is, '*jon painted his dog*' is interpreted as follows: there is a unique object named *jon*, which is an object of type Human, some object *d* which of type Dog, such that *d* actually exists, *jon* does OWN *d*, and *jon* did PAINT *d*. The point of the above example was to illustrate that the notion of intensional verbs can be captured in this simple formalism without the type lifting operation, particularly since an extensional interpretation might at times be implied even if an 'intensional' verb does not coexist with an extensional verb in the same context. As an illustrative example, let us as-sume $\text{PLAN}(x :: \text{Human}, y :: \text{Event})$; that is, that it always makes sense to say that some Human is planning (or did plan) something we call an Event. Consider now the following:

(28)

$[\![\,jon\ planned\ a\ trip\,]\!]$
$\Rightarrow (\exists^1 jon :: \text{Entity})(\exists e :: \text{Trip})$
$\qquad (\text{PLAN}(jon :: \text{Human}, e :: \text{Event}))$
$\Rightarrow (\exists^1 jon :: \text{Entity})(\exists e :: (\text{Trip} \bullet \text{Event}))(\text{PLAN}(jon, e))$
$\Rightarrow (\exists^1 jon :: \text{Entity})(\exists e :: \text{Trip})(\text{PLAN}(jon, e))$

That is, '*jon planned a trip*' simply states that a specific object that must be a Human has planned something we call a Trip (a trip that might not have actually happened[9]). Assuming $\text{LENGTHY}(e :: \text{Event}^{\square c})$, however, i.e., that LENGTHY is a property that is ordinarily said of an (existing) Event, then the interpretation of 'john planned the lengthy trip' should proceed as follows:

$[\![\,jon\ planned\ a\ lengthy\ trip\,]\!]$
$\Rightarrow (\exists^1 jon :: \text{Human})(\exists e :: \text{Trip})$
$\qquad (\text{PLAN}(jon, e :: \text{Event})) \wedge \text{LENGTHY}(e :: \text{Event}^{\square c}))$

Since $(\text{Trip} \bullet (\text{Event} \bullet \text{Event}^{\square c})) = (\text{Trip} \bullet \text{Event}^{\square c}) = \text{Trip}^{\square c}$ we finally get the following:

(29)    $[\![\,jon\ planned\ a\ lengthy\ trip\,]\!]$
$\qquad \Rightarrow (\exists^1 jon :: \text{Entity})(\exists e :: \text{Trip}^{\square c})$
$\qquad\qquad (\text{PLAN}(jon, e) \wedge \text{LENGTHY}(e))$
$\qquad \Rightarrow (\exists^1 jon :: \text{Entity})(\exists e :: \text{Trip})$
$\qquad\qquad (\text{PLAN}(jon, e) \wedge Exist(e) \wedge \text{LENGTHY}(e))$

That is, there is a specific Human named *jon* that has planned a Trip, a trip that actually exists, and a trip that was LENGTHY. Finally, it should be noted here that the trip in (29) was finally considered to be an existing Event due to other information contained in the same sentence. In general, however, this information can be contained in a larger discourse. For example, in interpreting '*John planned a trip. It was lengthy*' the resolution of 'it' would force a retraction of the types inferred in processing '*John planned a trip*', as the information that follows will 'bring down' the aforementioned Trip from abstract to actual existence (or, from mere being to concrete existence). This discourse level analysis is clearly beyond the scope of this paper, but readers interested in the computational details of such processes are referred to (van Deemter & Peters, 1996).

### 3.4   Metonymy and Copredication

In addition to so-called intensional verbs, our proposal seems to also appropriately handle other situations that, on the surface, seem to be addressing a different issue. For example, consider the following:

(30)    *Jon read the book and then he burned it.*

In Asher and Pustejovsky (2005) it is argued that this is an example of what they term copredication; which is the possibility of incompatible predicates to be applied to the same type of object. It is argued that in (30), for example, 'book' must have what is called a dot type, which is a complex structure that in a sense carries the 'informational content' sense (which is referenced when it is being read) as well as the 'physical object' sense (which is referenced when it is being burned). Elaborate machinery is then introduced to 'pick out' the right sense in the right context, and all in a well-typed compositional logic. But this approach presupposes that one can enumerate, a priori, all possible uses of the word 'book' in ordinary language[10]. Moreover, copredication seems to be a special case of metonymy, where the possible relations that could be implied are in fact much more constrained. An approach that can explain both notions, and hopefully without introducing much complexity into the logical form, should then be more desirable.

Let us first suggest the following:

(31)    a. $\text{READ}(x :: \text{Human}, y :: \text{Content})$
        b. $\text{BURN}(x :: \text{Human}, y :: \text{Physical})$

That is, we are assuming here that speakers of ordinary language understand 'read' and 'burn' as follows: it always makes sense to speak of a Human that read some Content, and of a Human that burned some Physical object. Consider now the following:

(32)    $[\![\,jon\ read\ a\ book\ and\ then\ he\ burned\ it\,]\!]$
$\qquad \Rightarrow (\exists^1 jon :: \text{Entity})(\exists b :: \text{Book})$
$\qquad\qquad (\text{READ}(jon :: \text{Human}, b :: \text{Content}))$
$\qquad\qquad \wedge \text{BURN}(jon :: \text{Human}, b :: \text{Physical}))$

The type unification of *jon* is straightforward, as the agent of BURN and READ are of the same type. Concerning $b$, a pair of type unifications $((\text{Book} \bullet \text{Physical}) \bullet \text{Content})$ must occur, resulting in the following:

(33)    $[\![\,jon\ read\ a\ book\ and\ then\ he\ burned\ it\,]\!]$
$\qquad \Rightarrow (\exists^1 jon :: \text{Entity})(\exists b :: (\text{Book} \bullet \text{Content}))$
$\qquad\qquad (\text{READ}(jon, b) \wedge \text{BURN}(jon, b)))$

Since no subsumption relation exists between Book and Content, the two variables are kept and a salient relation between them is introduced, resulting in the following:

(34)    $[\![\,jon\ read\ a\ book\ and\ then\ he\ burned\ it\,]\!]$
$\qquad \Rightarrow (\exists^1 jon :: \text{Entity})(\exists b :: \text{Book})(\exists c :: \text{Content})$
$\qquad\qquad (\mathbf{R}(b, c) \wedge \text{READ}(jon, c) \wedge \text{BURN}(jon, b))$

That is, there is some unique object of type Human (named *jon*), some Book $b$, some content $c$, such that $c$ is the Content of $b$, and such that *jon* read $c$ and burned $b$.

---

As in the case of copredication, type unifications introducing an additional variable and a salient relation occurs also in situations where we have what we refer to as metonymy. To illustrate, consider the following example:

(35) $[\![$*the ham sadnwich wants a beer*$]\!]$
$\Rightarrow (\exists^1 x :: \mathsf{HamSandwich})(\exists y :: \mathsf{Beer})$
$\quad (\mathrm{WANT}(x :: \mathsf{Human}, y :: \mathsf{Thing}))$
$\Rightarrow (\exists^1 x :: \mathsf{HamSandwich})(\exists y :: (\mathsf{Beer} \bullet \mathsf{Thing}))$
$\quad (\mathrm{WANT}(x :: \mathsf{Human}, y))$
$\Rightarrow (\exists^1 x :: \mathsf{HamSandwich})(\exists y :: \mathsf{Beer})$
$\quad (\mathrm{WANT}(x :: \mathsf{Human}, y))$

While the type unification between Beer and Thing is trivial, since $(\mathsf{Beer} \sqsubseteq \mathsf{Thing})$, the type unification involving the variable $x$ fails since there is no subsumption relationship between Human and HamSandwich. As argued above, in these situations both types are kept and a salient relation between them is introduced, as follows:

$[\![$*the ham sadnwich wants a beer*$]\!]$
$\Rightarrow (\exists^1 x :: \mathsf{HamSandwich})(\exists^1 z :: \mathsf{Human})(\exists y :: \mathsf{Beer})$
$\quad (\mathbf{R}(x, z) \wedge \mathrm{WANT}(z, y))$

where $\mathbf{R} = msr(\mathsf{Human}, \mathsf{Sandwich})$, i.e., where $\mathbf{R}$ is assumed to be some salient relation (e.g., EAT, ORDER, etc.) that exists between an object of type Human, and an object of type Sandwich (more on this below).

### 3.5 Types and Salient Relations

Thus far we have assumed the existence of a function $msr(\mathsf{s}, \mathsf{t})$ that returns, if it exists, the most salient relation $\mathbf{R}$ between two types $\mathsf{s}$ and $\mathsf{t}$. Before we discuss what this function might look like, we need to extend the notion of assigning ontological types to properties and relations slightly. Let us first reconsider (1), which is repeated below:

(36)    *Pass that car, will you.*
    a. *He is really annoying me.*
    b. *They are really annoying me.*

As discussed above we argue that 'he' in (36a) refers to 'the person driving [that] car' while 'they' in (36b) refers to 'the people riding in [that] car'. The question here is this: although there are many possible relations between a Person and a Car (e.g., DRIVE, RIDE, MANUFACTURE, DESIGN, MAKE, etc.) how is it that DRIVE is the one that most speakers assume in (36a), while RIDE is the one most speakers would assume in (36b)? Here's a plausible answer:

- DRIVE is more salient than RIDE, MANUFACTURE, DESIGN, MAKE, etc. since the other relations apply higher-up in the hierarchy; that is, the fact that we MAKE a Car, for example, is not due to Car, but to the fact that MAKE can be said of any Artifact and $(\mathsf{Car} \sqsubseteq \mathsf{Artifact})$.

- While DRIVE is a more salient relation between a Human and a Car than RIDE, most speakers of ordinary

English understand the DRIVE relation to hold between one Human and one Car (at a specific point in time), while RIDE is a relation that holds between many (several, or few!) people and one car. Thus, 'they' in (36b) fails to unify with DRIVE, and the next most salient relation must be picked up, which in this case is RIDE.

In other words, the type assignments of DRIVE and RIDE are understood by speakers of ordinary language as follows:

$\mathrm{DRIVE}(x :: \mathsf{Human}^1, y :: \mathsf{Car}^1)$
$\mathrm{RIDE}(x :: \mathsf{Human}^{1^+}, y :: \mathsf{Car}^1)$

With this background, let us now suggest how the function $msr(\mathsf{s}, \mathsf{t})$ that picks out the most salient relation $\mathbf{R}$ between two types $\mathsf{s}$ and $\mathsf{t}$ is computed.

We say $\mathrm{pap}(\mathsf{P}, \mathsf{t})$ when the property $\mathsf{P}$ applies to objects of type $\mathsf{t}$, and $\mathrm{rap}(\mathsf{R}, \mathsf{s}, \mathsf{t})$ when the relation r holds between objects of type $\mathsf{s}$ and objects of type $\mathsf{t}$. We define a list $\mathrm{lpap}(\mathsf{t})$ of all properties that apply to objects of type $\mathsf{t}$, and $\mathrm{lrap}(\mathsf{s}, \mathsf{t})$ of all relations that hold between objects of type $\mathsf{s}$ and objects of type $\mathsf{t}$, as follows:

(37)    $\mathrm{lpap}(\mathsf{t}) = [\mathsf{P} \mid \mathrm{pap}(\mathsf{P}, \mathsf{t})]$
    $\mathrm{lrap}(\mathsf{s}, \mathsf{t}) = [\langle \mathsf{R}, m, n \rangle \mid \mathrm{rap}(\mathsf{R}, \mathsf{s}^m, \mathsf{t}^n)]$

The lists (of lists) $\mathrm{lpap}^*(\mathsf{t})$ and $\mathrm{lrap}^*(\mathsf{s}, \mathsf{t})$ can now be inductively defined as follows:

(38)    $\mathrm{lpap}^*(\mathsf{Thing}) = [\,]$
    $\mathrm{lpap}^*(\mathsf{t}) = \mathrm{lpap}(\mathsf{t}) : \mathrm{lpap}^*(\sup(\mathsf{t}))$

    $\mathrm{lrap}^*(\mathsf{s}, \mathsf{Thing}) = [\,]$
    $\mathrm{lrap}^*(\mathsf{s}, \mathsf{t}) = \mathrm{lrap}(\mathsf{s}, \mathsf{t}) : \mathrm{lrap}^*(\mathsf{s}, \sup(\mathsf{t}))$

where $(e : s)$ is a list that results from attaching the object $e$ to the front of the (ordered) list $s$, and where $\sup(\mathsf{t})$ returns the immediate (and single!) parent of $\mathsf{t}$. Finally, we now define the function $\mathrm{msr}(\langle \mathsf{s}^m, \mathsf{t}^n \rangle)$ which returns most the salient relation between objects of type $\mathsf{s}$ and $\mathsf{t}$, with constraints $m$ and $n$, respectively, as follows:

$\mathrm{msr}(\langle \mathsf{s}^m, \mathsf{t}^n \rangle) = \mathtt{if}\ (\mathsf{s} \neq [\,])\ \mathtt{then}\ (\mathtt{head}\ \mathsf{s})\ \mathtt{else}\ \bot$
where
$\mathsf{s} = [\mathsf{R} \mid \langle \mathsf{R}, a, b \rangle \in \mathrm{lrap}^*(\mathsf{s}, \mathsf{t}) \wedge (a \geq m) \wedge (b \geq n)]$

Assuming now the ontological and logical concepts shown in figure 1, for example, then

$\mathrm{lpap}^*(\mathsf{Human})$
$= [[\mathrm{ARTICULATE}, \ldots], [\mathrm{HUNGRY}, \ldots], [\mathrm{HEAVY}, \ldots], [\mathrm{OLD}, \ldots], \ldots]$
$\mathrm{lrap}^*(\mathsf{Human}, \mathsf{Car})$
$= [[\langle \mathrm{DRIVE}, 1, 1 \rangle, \ldots], [\langle \mathrm{RIDE}, 1^+, 1 \rangle, \ldots], \ldots, \ldots]$

Since these lists are ordered, the degree to which a property or a relation is salient is inversely related to the position of the property or the relation in the list. Thus, for example, while a Human may DRIVE, RIDE, MAKE, BUY, SELL, BUILD, etc. a Car, DRIVE is a more salient relation between a Human and a Car than RIDE, which, in turn, is more salient than MANUFACTURE, MAKE, etc. Moreover, assuming the above sets we have

$$\texttt{msr}(\langle \mathsf{Human}^1, \mathsf{Car}^1 \rangle) = \text{DRIVE}$$
$$\texttt{msr}(\langle \mathsf{Human}^{1+}, \mathsf{Car}^1 \rangle) = \text{RIDE}$$

which essentially says DRIVE is the most salient relation in a context where we are speaking of a single Human and a single Car, and RIDE is the most salient relation between a number of people and a Car. Note now that 'they' in (36b) can be interpreted as follows:

$\llbracket$*They are annoying me*$\rrbracket$
$\Rightarrow (\exists they :: (\mathsf{Human}^{1+} \bullet \mathsf{Car}))(\exists me :: \mathsf{Human}^1)$
$\qquad (\text{ANNOYING}(they, me))$
$\Rightarrow (\exists they :: \mathsf{Human}^{1+})(\exists c :: \mathsf{Car})(\exists me :: \mathsf{Human}^1)$
$\qquad (\text{RIDING}(they, c) \wedge \text{ANNOYING}(they, me))$

It should be clear from the above that type unification and computing the most salient relation between two (ontological) types is what determines that *Jon* enjoyed 'reading' the book in (39a), and enjoyed 'watching' the movie in (39b).

(39) a. *Jon enjoyed the book.*
   b. *Jon enjoyed the movie.*

Note however that in addition to READ, an object of type Human may also WRITE, BUY, SELL, etc a Book. Similarly, in addition to WATCH, an object of type Human may also CRITICIZE, DIRECT, PRODUCE, etc. a Movie. Although this issue is beyond the scope of the current paper we simply note that picking out the most salient relation is still decidable due to tow differences between READ/WRITE and WATCH/DIRECT (or WATCH/PRODUCE): (i) the number of people that usually read a book (watch a movie) is much greater than the number of people that usually write a book (direct/produce) a movie, and saliency is inversely proportional to these numbers; and (ii) our ontology typically has a specific name for those who write a book (author), and those who direct (director) or produce (producer) a movie.

## 4   ONTOLOGICAL TYPES AND THE COPULAR

Consider the following sentences involving two different uses of the copular 'is':

(40) a. *William H. Bonney is Billy the Kid.*
   b. *Liz is famous.*

The copular 'is' in (40a) is usually referred to as the 'is of identity' while that in (40b) as the 'is of predication' and the standard first-order logic translation of the sentences in (40) is usually given by (41a) and (41b), respectively (using *whb* for *William H. Bonney* and *btk* for *Bill the Kid*):

(41) a. $whb = btk$
   b. $Famous(Liz)$

However, we argue that 'is' is not ambiguous but, like any other relation, it can occur in contexts in which an additional salient relation is implied, depending on the types of the objects involved. Thus, we have the following:

(42) $\llbracket whb \text{ is } btk \rrbracket$
$\Rightarrow (\exists^1 whb :: \mathsf{Human})(\exists^1 btk :: \mathsf{Human})(\mathbf{BE}(whb, btk))$
$\Rightarrow (\exists^1 whb :: \mathsf{Human})(\exists^1 btk :: \mathsf{Human})(\mathbf{EQ}(whb, btk))$

Note that since both objects are of the same type, **BE** in (42) is trivially translated into an equality. However, consider now the following:

(43) $\llbracket liz \text{ is } famous \rrbracket$
$\Rightarrow (\exists^1 liz :: \mathsf{Human})(\exists p :: \mathsf{Property})$
$\qquad (\text{FAME}(p) \wedge \mathbf{BE}(liz :: \mathsf{Human}, p :: \mathsf{Property}))$

As we have done thus far, since no subsumption relation exists between Human and Property, some salient relation must be introduced, where the most salient relation between an object $x$ and a property $y$ is $\mathbf{HAS}(x,y)$, meaning that $x$ has the property $y$:

$\llbracket liz \text{ is } famous \rrbracket$
$\Rightarrow (\exists^1 liz :: \mathsf{Human})$
$\qquad (\exists p :: \mathsf{Property})(\text{FAME}(p) \wedge \mathbf{HAS}(liz, p))$

Thus, saying that '*Liz is famous*' is saying that there is some unique object named *Liz*, an object of type Human, and some Property $p$, such that *Liz* has that property. A similar analysis yields the following interpretations:

(44)
a. $\llbracket$*aging is inevitable*$\rrbracket$
   $\Rightarrow (\exists^1 x :: \mathsf{Process})(\exists^1 y :: \mathsf{Property})$
   $\qquad (\text{AGING}(x) \wedge \text{INEVITABILITY}(y) \wedge \mathbf{HAS}(x, y))$
b. $\llbracket$*fame is desirable*$\rrbracket$
   $\Rightarrow (\exists^1 x :: \mathsf{Property})(\exists^1 y :: \mathsf{Property})$
   $\qquad (\text{FAME}(x) \wedge \text{DESIRABILITY}(y) \wedge \mathbf{HAS}(x, y))$
c. $\llbracket$*sheba is dead*$\rrbracket$
   $\Rightarrow (\exists^1 x :: \mathsf{Human})(\exists^1 y :: \mathsf{State})(\text{DEATH}(y) \wedge \mathbf{IN}(x, y))$
d. $\llbracket$*jon is aging*$\rrbracket$
   $\Rightarrow (\exists^1 jon :: \mathsf{Human})(\exists^1 y :: \mathsf{Process})$
   $\qquad (\text{AGING}(y) \wedge \mathbf{GT}(x, y))$

That is, the Process of AGING has the Property of being inevitable (44a); the Property FAME has the (other) Property of being DESRIBALE (44b); *sheba* is in a (physical) State called DEATH (44c); and, finally, *jon* is going through (**GT**) a Process called AGING (44d). Finally, consider the following well-known example (due, we believe, to Barbara Partee):

(45) a. *The temperature is 90.*
   b. *The temperature is rising.*
   c. *90 is rising.*

It has been argued that such sentences require an intensional treatment since a purely extensional treatment would make

(54a) and (54b) erroneously entail (45c). However, we believe that the embedding of ontological types into the properties and relations yields the correct entailments without the need for complex higher-order intensional formalisms. Consider the following:

$$[\![\textit{the temperature is } 90]\!]$$
$$\Rightarrow (\exists^! x :: \mathsf{Temperature})(\exists^! y :: \mathsf{Measure})$$
$$(\mathrm{VALUE}(y, 90)$$
$$\wedge \, \mathbf{BE}(x :: \mathsf{Temperature}, y :: \mathsf{Measure}))$$

Since no subsumption relation exits between an object of type Temperature and an object of type Measure, the type unification in $\mathbf{BE}(x, y)$ should result in a salient relation between the two types, as follows;

(46) $[\![\textit{the temperature is } 90]\!]$
$$\Rightarrow (\exists^! x :: \mathsf{Temperature})(\exists^! y :: \mathsf{Measure})$$
$$(\mathrm{VALUE}(y, 90) \wedge \mathbf{HAS}(x, y))$$

On the other hand, consider now the following:

(47) $[\![\textit{the temperature is rising}]\!]$
$$\Rightarrow (\exists^! x :: \mathsf{Temperature})(\exists^! y :: \mathsf{Process})(\mathbf{BE}(x, y))$$

Again, as no subsumption relation exists between an object of type Temperature and an object of type Process, some salient relation between the two is introduced. However, in this case the salient relation is quite different; in particular, the relation is that of $x$-going-through the State $y$:

(48) $[\![\textit{the temperature is rising}]\!]$
$$\Rightarrow (\exists^! x :: \mathsf{Temperature})(\exists^! y :: \mathsf{Process})$$
$$(\mathrm{RISING}(y) \wedge \mathbf{GT}(x, y))$$

Note now that (46) and (48) yield the following, which essentially says that 'the temperature is 90 and it is rising':

$$(\exists^! x :: \mathsf{Temperature})(\exists^! y :: \mathsf{Measure})(\exists z :: \mathsf{Process})$$
$$(\mathrm{RISING}(z) \wedge \mathrm{VALUE}(y, 90)$$
$$\wedge \mathbf{HAS}(x, y) \wedge \mathbf{GT}(x, z)))$$

Finally, note that uncovering the ontological commitments implied by the sentences in (45a) and (54b) will not result in the erroneous entailment of (45c).

Contrary to the situation in (45), however, uncovering the ontological commitments implied by some sentences should sometimes admit some valid entailments. For example, consider the following:

(49) a. *exercising is wise.*
     b. *jon is exercising.*
     c. *jon is wise.*

Clearly, (49a) and (49b) should entail (49c), although one can hardly think of attributing the property WISE to an Activity (EXERCISING). Let us see how we might explain this argument. We start with the simplest:

(50) $[\![\textit{jon is exercising}]\!]$
$$\Rightarrow (\exists^! jon :: \mathsf{Human})(\exists^! act :: \mathsf{Activity})$$
$$(\mathrm{EXERCISING}(act) \wedge \mathrm{AGENT}(act, jon))$$

Let us now consider the following:

(51) $[\![\textit{exercising is wise}]\!]$
$$\Rightarrow (\forall a :: \mathsf{Activity})(\mathrm{EXERCISING}(a)$$
$$\supset (\exists^! p :: \mathsf{Property})(\mathrm{WISDOM}(p)$$
$$\wedge \mathbf{HAS}(a :: \mathsf{Human}, p))$$

That is, any exercising Activity has a property, namely WISDOM, which is a property that ordinarily an object of type Human has. Note, however, that a type unification for the variable $a$ must now occur:

(52) $[\![\textit{jon is exercising}]\!]$
$$\Rightarrow (\forall a :: (\mathsf{Activity} \bullet \mathsf{Human}))(\mathrm{EXERCISING}(a)$$
$$\supset (\exists^! p :: \mathsf{Property})$$
$$(\mathrm{WISDOM}(p) \wedge \mathbf{HAS}(a, p))$$

The most salient relation between a Human and an Activity is that of agency – that is, a human is typically the AGENT of an activity:

(53) $[\![\textit{jon is exercising}]\!]$
$$\Rightarrow (\forall a :: \mathsf{Activity})(\forall x :: \mathsf{Human})(\mathrm{EXERCISING}(a)$$
$$\wedge \mathrm{AGENT}(a, x) \supset (\exists^! p :: \mathsf{Property})$$
$$(\mathrm{WISDOM}(p) \wedge \mathbf{HAS}(a, p))$$

Essentially, therefore, we get the following: any human $x$ has the property of being WISE whenever $x$ is the agent of an exercising activity. Note now that (50), (53) and modes ponens results in the following, which is the meaning of '*jon is wise*':

$$(\exists^! jon :: \mathsf{Human})$$
$$(\exists^! p :: \mathsf{Property})(\mathrm{WISDOM}(p) \wedge \mathbf{HAS}(x, p))$$

Finally, note that the inference in (49) was proven valid only after uncovering the missing text, since '*exercising is wise*' was essentially interpreted as '[any human that performs the activity of] *exercising is wise*'.

## 5   CONCLUDING REMARKS

If the main business of semantics is to explain how linguistic constructs relate to the world, then semantic analysis of natural language text is, indirectly, an attempt at uncovering the semiotic ontology of commonsense knowledge, and particularly the background knowledge that seems to be implicit in all that we say in our everyday discourse. While this intimate relationship between language and the world is generally accepted, semantics (in all its paradigms) has traditionally proceeded in one direction: by first stipulating an assumed set of ontological commitments followed by some machinery that is supposed to, somehow, model meanings in terms of that stipulated structure of reality.

With the gross mismatch between the trivial ontological commitments of our semantic formalisms and the reality of the world these formalisms purport to represent, it is not surprising therefore that challenges in the semantics of natural language are rampant. However, as correctly observed by Hobbs (1985), semantics could become nearly trivial if it was grounded in an ontological structure that is "isomorphic to the way we talk about the world". The obvious question however is 'how does one arrive at this ontological structure that implicitly underlies all that we say in everyday discourse?' One plausible answer is the (seemingly circular) suggestion that the semantic analysis of natural language should itself be used to uncover this structure. In this regard we strongly agree with Dummett (1991) who states:

> We must not try to resolve the metaphysical questions first, and then construct a meaning-theory in light of the answers. We should investigate how our language actually functions, and how we can construct a workable systematic description of how it functions; the answers to those questions will then determine the answers to the metaphysical ones.

What this suggests, and correctly so, in our opinion, is that in our effort to understand the complex and intimate relationship between ordinary language and everyday commonsense knowledge, one could, as also suggested in (Bateman, 1995), "use language as a tool for uncovering the semiotic ontology of commonsense" since ordinary language is the best known theory we have of everyday knowledge. To avoid this seeming circularity (in wanting this ontological structure that would trivialize semantics; while at the same time suggesting that semantic analysis should itself be used as a guide to uncovering this ontological structure), we suggested here performing semantic analysis from the ground up, assuming a minimal (almost a trivial and basic) ontology, in the hope of building up the ontology as we go guided by the results of the semantic analysis. The advantages of this approach are: (*i*) the ontology thus constructed as a result of this process would not be invented, as is the case in most approaches to ontology (e.g., Lenat, & Guha (1990); Guarino (1995); and Sowa (1995)), but would instead be discovered from what is in fact implicitly assumed in our use of language in everyday discourse; (*ii*) the semantics of several natural language phenomena should as a result become trivial, since the semantic analysis was itself the source of the underlying knowledge structures (in a sense, the semantics would have been done before we even started!)

Throughout this paper we have tried to demonstrate that a number of challenges in the semantics of natural language can be easily tackled if semantics is grounded in a strongly-typed ontology that reflects our commonsense view of the world and the way we talk about it in ordinary language. Our ultimate goal, however, is the systematic discovery of this ontological structure, and, as also argued in Saba (2007), it is the systematic investigation of how ordinary language is used in everyday discourse that will help us discover (as opposed to invent) the ontological structure that seems to underlie all what we say in our everyday discourse.

## ACKNOWLEDGEMENT

While any remaining errors and/or shortcomings are our own, the work presented here has benefited from the valuable feedback of the reviewers and attendees of the *13th Portuguese Conference on Artificial Intelligence* (EPIA 2007), as well as those of Romeo Issa of Carleton University and those of Dr. Graham Katz and his students at Georgetown University.